\documentclass[final]{cvpr}
\usepackage{times}
\usepackage{epsfig}
\usepackage{graphicx}
\usepackage{amsmath}
\usepackage{amssymb}
\usepackage{multirow}
\usepackage{color}
\usepackage{color}

\DeclareMathOperator*{\argmin}{argmin}
\usepackage{algorithm}  
\usepackage{algorithmic} 
\usepackage{bm}
\usepackage{booktabs}
\usepackage{array}
\usepackage{bbold}
\newcolumntype{C}[1]{>{\centering}p{#1}}
\usepackage{lipsum}
\newcommand\nmfootnote[1]{%
  \begingroup
  \renewcommand\thefootnote{}\footnote{#1}%
  \addtocounter{footnote}{-1}%
  \endgroup
}

\usepackage[pagebackref=true,breaklinks=true,colorlinks,bookmarks=false]{hyperref}

\def\cvprPaperID{453} 
\def\confYear{CVPR 2021}

\begin{document}

\title{Multiple Instance Active Learning for Object Detection}

\author{
Tianning Yuan\textsuperscript{$\dag$}, Fang Wan\textsuperscript{$\dag$\footnotemark[1]}, Mengying Fu\textsuperscript{$\dag$},\\
Jianzhuang Liu\textsuperscript{$\ddag$}, Songcen Xu\textsuperscript{$\ddag$}, Xiangyang Ji\textsuperscript{$\S$} and Qixiang Ye\textsuperscript{$\dag$\footnotemark[1]}\\
\\
\\
\textsuperscript{$\dag$}University of Chinese Academy of Sciences, Beijing, China\\
\textsuperscript{$\ddag$}Noah’s Ark Lab, Huawei Technologies, Shenzhen, China. \textsuperscript{$\S$}Tsinghua University, Beijing, China\\
\tt\small \{yuantianning19,fumengying19\}@mails.ucas.ac.cn, \{wanfang,qxye\}@ucas.ac.cn\\
\tt\small \{liu.jianzhuang,xusongcen\}@huawei.com, xyji@tsinghua.edu.cn

}

\maketitle

\nmfootnote{\textsuperscript*Corresponding Authors.} 

\begin{abstract}

Despite the substantial progress of active learning for image recognition, there still lacks an instance-level active learning method specified for object detection. In this paper, we propose Multiple Instance Active Object Detection (MI-AOD), to select the most informative images for detector training by observing instance-level uncertainty. MI-AOD defines an instance uncertainty learning module, which leverages the discrepancy of two adversarial instance classifiers trained on the labeled set to predict instance uncertainty of the unlabeled set. MI-AOD treats unlabeled images as instance bags and feature anchors in images as instances, and estimates the image uncertainty by re-weighting instances in a multiple instance learning (MIL) fashion. Iterative instance uncertainty learning and re-weighting facilitate suppressing noisy instances, toward bridging the gap between instance uncertainty and image-level uncertainty. Experiments validate that MI-AOD sets a solid baseline for instance-level active learning. On commonly used object detection datasets, MI-AOD outperforms state-of-the-art methods with significant margins, particularly when the labeled sets are small. 
Code is available at \href{https://github.com/yuantn/MI-AOD}{https://github.com/yuantn/MI-AOD}.

\end{abstract}

\begin{figure}[t]
    \centering
    \includegraphics[width=1\linewidth]{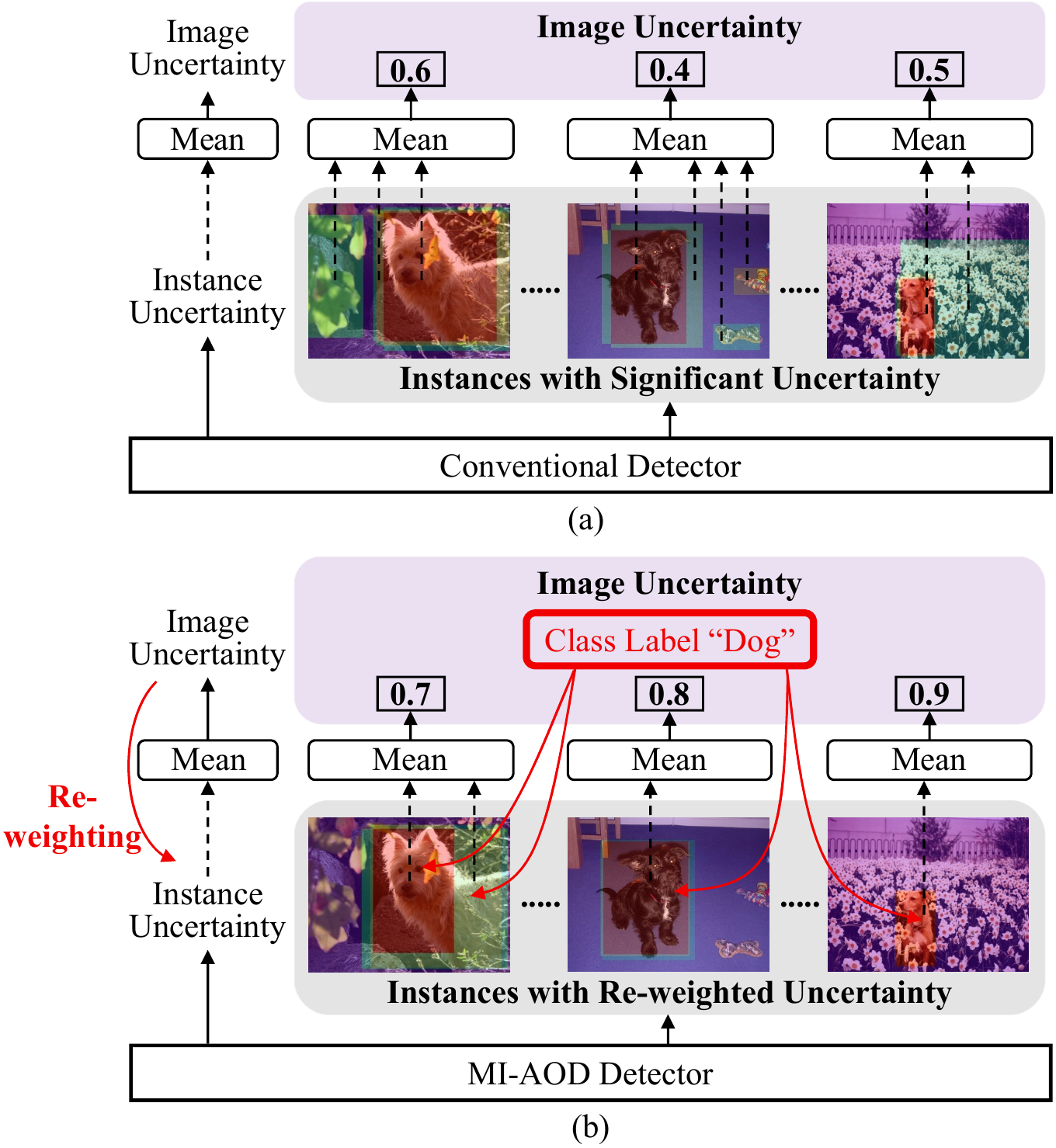}
    \caption{Comparison of active object detection methods. (a) Conventional methods compute image uncertainty by simply averaging instance uncertainties, ignoring interference from a large number of background instances. (b) Our MI-AOD leverages uncertainty re-weighting via multiple instance learning to filter out interfering instances while bridging the gap between instance uncertainty and image uncertainty. (Best viewed in color)
    }
    \label{fig:motivation}
\end{figure}

\section{Introduction}

The key idea of active learning is that a machine learning algorithm can achieve better performance with fewer training samples if it is allowed to select which to learn. Despite the rapid progress of the methods with less supervision~\cite{Lbh2021TIP,Lbh2021CVPR}, \eg, weak supervision and semi-supervision, active learning remains the cornerstone of many practical applications for its simplicity and higher performance.

In the computer vision area, active learning has been widely explored for image classification (active image classification) by empirically generalizing the model trained on the labeled set to the unlabeled set~\cite{DeepBayes17, CoreSet18, SelfPaced18,CostEffective17,PowerEnsem18,OpenSet2019,SelfPaced2020,UncertaintyGraph2020,VAAL19}.
Uncertainty-based methods define various metrics for selecting informative images and adapting trained models to the unlabeled set~\cite{DeepBayes17}. Distribution-based approaches \cite{CoreSet18, CDAL20} aim at estimating the layout of unlabeled images to select samples of large diversity. Expected model change methods~\cite{SelectInfluent14, ActiveContinuous16} find out samples that can cause the greatest change of model parameters or the largest loss~\cite{LearningLoss19}.

Despite the substantial progress, there still lacks an instance-level active learning method specified for object detection (\ie, active object detection~\cite{LearningLoss19, SRAAL20, PedesDetect19}), where the instance denotes the object proposal in an image.
The objective goal of active object detection is to select the most informative images for detector training. However, the recent methods tackled it by simply summarizing/averaging instance/pixel uncertainty as image uncertainty and unfortunately ignored the large imbalance of negative instances in object detection, which causes significant noisy instances in the background and interferes with the learning of image uncertainty, Fig.~\ref{fig:motivation}(a).
The noisy instances also cause the inconsistency between image and instance uncertainty, which hinders selecting informative images.

%
In this paper, we propose a Multiple Instance Active Object Detection (MI-AOD) approach, Fig.~\ref{fig:motivation}(b), and target at selecting informative images from the unlabeled set by learning and re-weighting instance uncertainty with discrepancy learning and multiple instance learning (MIL). To learn the instance-level uncertainty, MI-AOD first defines an instance uncertainty learning (IUL) module, which leverages two adversarial instance classifiers plugged atop the detection network (\eg, a feature pyramid network) to learn the uncertainty of unlabeled instances. Maximizing the prediction discrepancy of two instance classifiers predicts instance uncertainty while minimizing classifiers’ discrepancy drives learning features to reduce the distribution bias between the labeled and unlabeled instances. 

To establish the relationship between instance uncertainty and image uncertainty, MI-AOD incorporates a MIL module, which is in parallel with the instance classifiers. MIL treats each unlabeled image as an instance bag and performs instance uncertainty re-weighting (IUR) by evaluating instance appearance consistency across images. During MIL, the instance uncertainty and image uncertainty are forced to be consistently driven by a classification loss defined on image class labels (or pseudo-labels). Optimizing the image-level classification loss facilitates suppressing the noisy instances while highlighting truly representative ones. Iterative instance uncertainty learning and instance uncertainty re-weighting bridge the gap between instance-level observation and image-level evaluation, towards selecting the most informative images for detector training.

The contributions of this paper include:

(1) We propose Multiple Instance Active Object Detection (MI-AOD), establishing a solid baseline to model the relationship between the instance uncertainty and image uncertainty for informative image selection. 

(2) We design instance uncertainty learning (IUL) and instance uncertainty re-weighting (IUR) modules, providing effective approaches to highlight informative instances while filtering out noisy ones in object detection.

(3) We apply MI-AOD to object detection on commonly used datasets, improving state-of-the-art methods with significant margins.

\section{Related Work}
\subsection{Active Learning}

\textbf{Uncertainty-based Methods.} Uncertainty is the most popular metric to select samples for active learning~\cite{Survey12}, which can be defined as the posterior probability of a predicted class~\cite{TextClassifier94, Heterogeneous94}, or the margin between the posterior probabilities of the first and the second predicted class~\cite{MultiClass09, MarginBased06}. It can also be defined upon entropy~\cite{SequenceLabel08, LatentStruct13, MultiClass09} to measure the variance of unlabeled samples. 
The expected model change methods~\cite{ErrorReduction01,MultiInstance07} utilized the present model to estimate the expected gradient or prediction changes~\cite{SelectInfluent14, ActiveContinuous16} for sample selection. MIL-based methods~\cite{MultiInstance07, MultiInstanceLabel17, IncorpDiver17, BagLevelAggre19} selected informative images by discovering representative instances. However, they are designed for image classification and not applicable to object detection due to the challenging aspect of crowded and noisy instances~\cite{MinEntropy2019,CMIL2019}.

\begin{figure*}[t]
    \centering
    \includegraphics[width=1\linewidth]{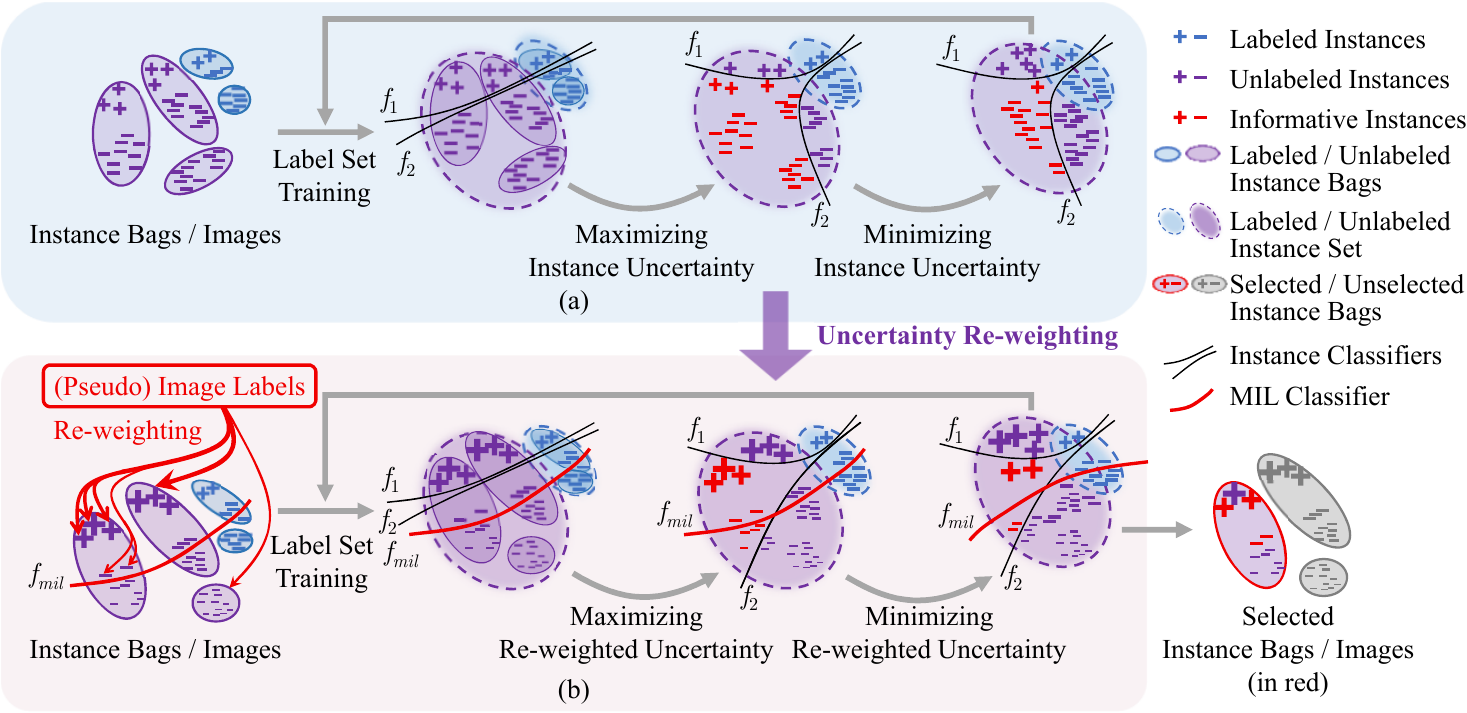}
    \caption{MI-AOD illustration. (a) Instance uncertainty learning (IUL) utilizing two adversarial classifiers. (b) Instance uncertainty re-weighting (IUR) using multiple instance learning. Bigger symbols (``$+$'' and ``$-$'') indicate larger weights. (Best viewed in color)}
    \label{fig:illustration}
\end{figure*}

\textbf{Distribution-based Methods.} These methods select diverse samples by estimating the distribution of unlabeled samples. Clusters~\cite{PreCluster04} were applied to build the unlabeled sample distribution while discrete optimization methods~\cite{MatrixPartition10,ConvexOptim13,MultiClassUncerDiver15} were employed to perform sample selection. Considering the distances to the surrounding samples, the context-aware methods~\cite{ContextAware15, HierarchicalSubquery14} selected the samples that can represent the global sample distribution. Core-set~\cite{CoreSet18} defined active learning as core-set selection, \ie, choosing a set of points such that a model learned on the labeled set can capture the diversity of the unlabeled samples. 

In the deep learning era, active learning methods remain falling into the uncertainty-based or distribution-based routines~\cite{SelfPaced18,CostEffective17,PowerEnsem18}. Sophisticated methods extended active learning to the open sets~\cite{OpenSet2019}, or combined it with self-paced learning~\cite{SelfPaced2020}. Nevertheless, it remains questionable whether the intermediate feature representation is effective for sample selection. The learning loss method~\cite{LearningLoss19} can be categorized as either uncertainty-based or distribution-based. By introducing a module to predict the ``loss'' of the unlabeled samples, it estimates sample uncertainty and selects samples with large ``loss'' like hard negative mining.

\subsection{Active Learning for Object Detection}
Despite the substantial progress of active learning, few methods are specified for active object detection, which faces complex instance distributions in the same images and are more challenging than active image classification. By simply sorting the loss predictions of instances to evaluate the image uncertainty, the learning loss method~\cite{LearningLoss19} specified for image classification was directly applied to object detection. The image-level uncertainty can also be estimated by the uncertainty of a lot of background pixels~\cite{PedesDetect19}. CDAL~\cite{CDAL20} introduced spatial context to active detection and selected diverse samples according to the distances to the labeled set. Existing approaches simply used instance-/pixel-level observations to represent the image-level uncertainty. There still lacks a systematic method to learn the image uncertainty by leveraging instance-level models~\cite{FreeAnchor2019,LTM2021}.


\begin{figure*}[t]
    \centering
    \includegraphics[width=1\textwidth]{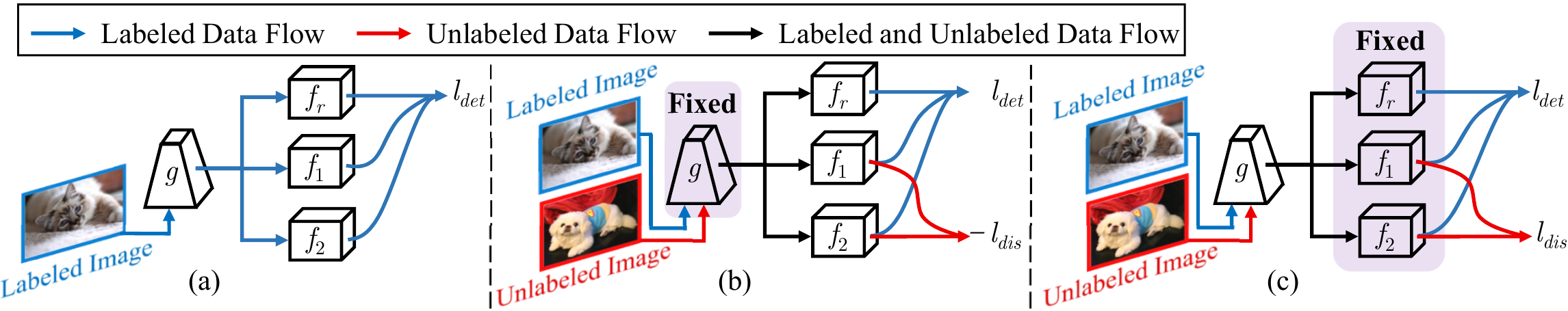}
    \caption{Network architecture for instance uncertainty learning. (a) Label set training. (b) Maximizing instance uncertainty by maximizing classifier prediction discrepancy. (c) Minimizing instance uncertainty by minimizing classifier prediction discrepancy.}
    \label{fig:flowchart-iul}
\end{figure*}

\section{The Proposed Approach}

\subsection{Overview}
For active object detection, a small set of images ${\cal X}^0_L$ (the labeled set) with instance labels ${\cal Y}^0_L$ and a large set of images ${\cal X}^0_U$ (the unlabeled set) without labels are given. For each image, the label consists of bounding boxes ($y_x^{loc}$) and categories ($y_x^{cls}$) for objects of interest. A detection model $M_0$ is firstly initialized by using the labeled set $\{{\cal X}^0_L, {\cal Y}^0_L\}$. With the initialized model $M_0$, active learning targets at selecting a set of images ${\cal X}_S^0$ from ${\cal X}^0_U$ to be manually labeled and merging them with ${\cal X}^0_L$ for a new labeled set ${\cal X}^1_L$, \ie, ${\cal X}^1_L = {\cal X}^0_L \cup {\cal X}^0_S$. The selected image set ${\cal X}_S^0$ should be the most informative, \ie, can improve the detection performance as much as possible. Based on the updated labeled set ${\cal X}^1_L$, the task model is retrained and updated to $M_1$. The model training and sample selection repeat some cycles until the size of labeled set reaches the annotation budget. 

Considering the large number\footnote{For example, the RetinaNet detector~\cite{RetinaNet20} produces $\sim$100k of anchors (instances) for an image.} of instances in each image, there are two key problems for active object detection: (1) how to evaluate the uncertainty of the unlabeled instances using the detector trained on the labeled set; (2) how to precisely estimate the image uncertainty while filtering out noisy instances. MI-AOD handles these two problems by introducing two learning modules respectively. For the first problem, MI-AOD incorporates instance uncertainty learning, with the aim of highlighting informative instances in the unlabeled set, as well as aligning the distributions of the labeled and unlabeled set, Fig.~\ref{fig:illustration}(a). It is motivated by the fact that most active learning methods remain simply generalizing the models trained on the labeled set to the unlabeled set. This is problematic when there is a distribution bias between the two sets~\cite{SelfSupervised2020}. For the second problem, MI-AOD introduces MIL to both the labeled and unlabeled set to estimate the image uncertainty by re-weighting the instance uncertainty. This is done by treating each image as an instance bag while re-weighting the instance uncertainty under the supervision of the image classification loss. Optimizing the image classification loss facilitates highlighting truly representative instances belonging to the same object classes while suppressing the noisy ones, Fig.~\ref{fig:illustration}(b). 

\subsection{Instance Uncertainty Learning}

\textbf{Label Set Training.} Using the RetinaNet as the baseline~\cite{RetinaNet20}, we construct a detector with two discrepant instance classifiers (${f_1}$ and ${f_2}$) and a bounding box regressor ($f_r$), Fig.~\ref{fig:flowchart-iul}(a). We utilize the prediction discrepancy between the two instance classifiers to learn the instance uncertainty on the unlabeled set. Let $g$ denote the feature extractor parameterized by $\theta_g$. The discrepant classifiers are parameterized by $\theta_{f_1}$ and $\theta_{f_2}$ and the regressor by $\theta_{f_r}$. $\Theta = \{\theta_{f_1}, \theta_{f_2}, \theta_{f_r}, \theta_{g}\}$ denotes the set of all parameters, where $\theta_{f_1}$ and $\theta_{f_2}$ are independently initialized.

In object detection, each image $x$ can be represented by multiple instances $ \{x_i, i=1,...,N\}$ corresponding to the feature anchors on the feature map~\cite{RetinaNet20}. $N$ is the number of the instances in image $x$. $\{y_{i}, i=1,...,N\}$ denote the labels for the instances. Given the labeled set, a detection model is trained by optimizing the detection loss, as
\begin{equation}
    \label{eq_detection}
    \begin{split}
        \argmin_{\Theta} {l}_{det}(x) =
        \sum_i \Big( & {\rm FL}(\hat{y}_i^{f_1}, {{y}_i^{cls}})
         + {\rm FL}(\hat{y}_i^{f_2},{y}_i^{cls})\\ 
         &+ {\rm SmoothL1}(\hat{y}_i^{f_r}, {y}_i^{loc})\Big),
    \end{split}
\end{equation}
where ${\rm FL}(\cdot)$ is the focal loss function for instance classification and ${\rm SmoothL1}(\cdot)$ is the smooth L1 loss function for bounding box regression~\cite{RetinaNet20}. $\hat{y}_i^{f_1} = f_1(g({x}_i))$, $\hat{y}_i^{f_2} = f_2(g({x}_i))$ and $\hat{y}_{i}^{f_r} = f_r(g(x_i))$ denote the prediction results (classification and localization) for the instances. ${y}_i^{cls}$ and ${y}_i^{loc}$ denote the ground-truth class label and bounding box label, respectively. 

\begin{figure*}[t]
    \centering
    \includegraphics[width=1\textwidth]{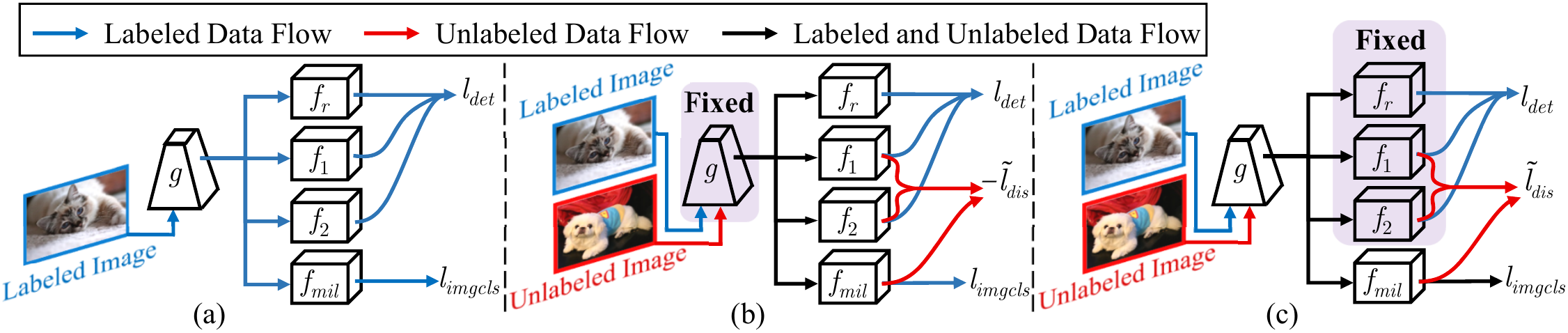}
    \caption{Network architecture for instance uncertainty re-weighting. (a) Label set training. (b) Re-weighting and maximizing instance uncertainty. (c) Re-weighting while minimizing instance uncertainty.}
    \label{fig:flowchart-iur}
\end{figure*}

\textbf{Maximizing Instance Uncertainty.} Before the labeled set can precisely represent the unlabeled set, there exists a distribution bias between the labeled and unlabeled set, especially when the labeled set is small. The informative instances are in the biased distribution area.
To find them out, $f_1$ and $f_2$ are designed as the adversarial instance classifiers with larger prediction discrepancy on the instances close to the boundary, Fig.~\ref{fig:illustration}(a). The instance uncertainty is defined as the prediction discrepancy of $f_1$ and $f_2$.

To find out the most informative instances, it requires to fine-tune the network and maximize the prediction discrepancy of the adversarial classifiers, Fig.~\ref{fig:flowchart-iul}(b). In this procedure, $\theta_g$ is fixed so that the distributions of both the labeled and unlabeled instances are fixed. $\theta_{f_1}$ and $\theta_{f_2}$ are fine-tuned on the unlabeled set to maximize the prediction discrepancies for all instances. At the same time, it requires to preserve the detection performance on the labeled set. This is fulfilled by optimizing the following loss function, as 
\begin{equation}
    \label{eq_max_discrepancy}
    \begin{split}
    \argmin_{\Theta \setminus \theta_{g}}{\cal L}_{max} &= \sum_{x\in {\cal X}_L}l_{det}(x) - 
    \sum_{x\in {\cal X}_U} \lambda \cdot l_{dis}(x),
    \end{split}
\end{equation}
where 
\begin{equation}
    \label{eq_max_discrepancy_l_dis}
    l_{dis}(x) = \sum_i |{\hat y_{i}^{f_1}} - {\hat y_{i}^{f_2}}|
\end{equation}
denotes the prediction discrepancy loss. ${\hat y_{i}^{f_1}}, {\hat y_{i}^{f_2}}\in\mathbb{R}^{1\times C}$ are the instance classification predictions of the two classifiers for the $i$-th instance in image $x$, where $C$ is the number of object classes in the dataset, and $\lambda$ is a regularization hyper-parameter determined by experiment. As shown in Fig.~\ref{fig:illustration}(a), the informative instances with different predictions by the adversarial classifiers tend to have larger prediction discrepancy and larger uncertainty.

\textbf{Minimizing Instance Uncertainty.} After maximizing the prediction discrepancy, we further propose to minimize the prediction discrepancy to align the distributions of the labeled and unlabeled instances, Fig.~\ref{fig:flowchart-iul}(c).
In this procedure, the classifier parameters $\theta_{f_1}$ and $\theta_{f_2}$ are fixed, while the parameters $\theta_g$ of the feature extractor are optimized by minimizing the prediction discrepancy loss, as 
\begin{equation}
    \label{eq_min_discrepancy}
    \begin{split}
        \argmin_{\theta_g}{\cal L}_{min} & = \sum_{x\in {\cal X}_L}l_{det}(x) + 
        \sum_{x\in {\cal X}_U} \lambda \cdot l_{dis}(x).
    \end{split}
\end{equation}
By minimizing the prediction discrepancy, the distribution bias between the labeled set and the unlabeled set is minimized and their features are aligned as much as possible. 

In each active learning cycle, the max-min prediction discrepancy procedures repeat several times so that the instance uncertainty is learned and the instance distributions of the labeled and unlabeled set are progressively aligned. This actually defines an unsupervised learning procedure, which leverages the information (\ie, prediction discrepancy) of the unlabeled set to improve the detection model. 

\subsection{Instance Uncertainty Re-weighting}
With instance uncertainty learning, the informative instances are highlighted. However, as there is a lot of instances ($\sim$100k) in each image, the instance uncertainty may be not consistent with the image uncertainty. Some instances of high uncertainty are simply background noise or hard negatives for the detector. We thereby introduce an MIL procedure to bridge the gap between the instance-level and image-level uncertainty by filtering out noisy instances.

\textbf{Multiple Instance Learning.} MIL treats each image as an instance bag and utilizes the instance classification predictions to estimate the bag labels. In turn, it re-weights the instance uncertainty scores by minimizing the image classification loss. This actually defines an Expectation-Maximization procedure~\cite{andrews2002MIL, Bilen2016Weakly} to re-weight instance uncertainty across bags while filtering out noisy instances. 

Specifically, we add an MIL classifier $f_{mil}$ parameterized by $\theta_{f_{mil}}$ in parallel with the instance classifiers, Fig.~\ref{fig:flowchart-iur}. The image classification score $\hat{y}^{cls}_{i,c}$ for multiple instances in an image is calculated as
\begin{equation}
    \label{eq_mil_score}
    \begin{split}
        \hat{y}^{cls}_{i,c} & = \frac{\exp(\hat{y}^{f_{mil}}_{i,c})}{\sum_c{\exp(\hat{y}^{f_{mil}}_{i,c})}} \cdot 
        \frac{\exp {\Big(}(\hat{{y}}^{f_1}_{i,c} + \hat{{y}}^{f_2}_{i,c})/2{\Big)}}{\sum_i \exp {\Big(}(\hat{{y}}^{f_1}_{i,c} + \hat{{y}}^{f_2}_{i,c})/2{\Big)}},
    \end{split}
\end{equation}
where ${\hat{y}^{f_{mil}}} = f_{mil}(g(x))$ is an $N\times C$ score matrix, and $\hat{y}^{f_{mil}}_{i,c}$ is the element in $\hat{y}^{f_{mil}}$ indicating the score of the $i$-th instance for class $c$. According to Eq.~(\ref{eq_mil_score}), the image classification score $\hat{y}^{cls}_{i,c}$ is large only when $x_i$ belongs to class $c$ (the first term in Eq.~(\ref{eq_mil_score})) and its instance classification scores $\hat{{y}}^{f_1}_{i,c}$ and $\hat{{y}}^{f_2}_{i,c}$ are significantly larger than those of others (the second term in Eq.~(\ref{eq_mil_score})). 

Considering that the image classification scores of the instances from other classes/backgrounds are small, the image classification loss $l_{imgcls}$ is defined as
\begin{equation}
    \label{eq_mil_loss}
    \begin{split}
        l_{imgcls}(x) 
        = - \sum_c {\Big(} & {y^{cls}_c}\log\sum_i \hat{y}^{cls}_{i,c} \\
                  + & (1-{y^{cls}_c})\log (1-\sum_i \hat{y}^{cls}_{i,c}){\Big)},
    \end{split}
\end{equation}
where $y^{cls}_c \in \{0, 1\}$ denotes the image class label, which can be directly obtained from the instance class label $y_i^{cls}$ in the labeled set.
Optimizing Eq.~(\ref{eq_mil_loss}) drives the MIL classifier to activate instances with large MIL score ($\hat{y}^{f_{mil}}_{i,c}$) and large classification outputs ($\hat{{y}}^{f_1}_{i,c} + \hat{{y}}^{f_2}_{i,c}$). 
The instances with small MIL scores will be suppressed as background. The image classification loss is firstly applied in the label set training to get the initial model, and then used to re-weight the instance uncertainty in the unlabeled set.

\begin{figure*}[ht!]
    \centering
    \includegraphics[width=1\textwidth]{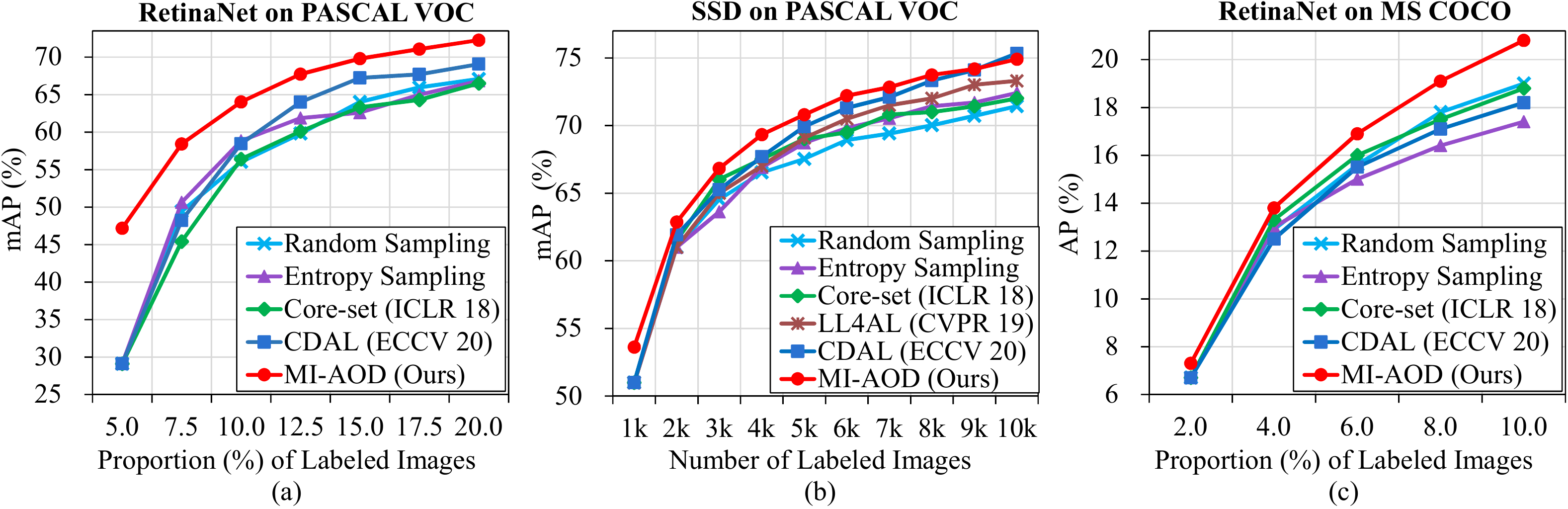}
    \caption{Performance comparison of active object detection methods. (a) On PASCAL VOC using RetinaNet. (b) On PASCAL VOC using SSD. (c) On MS COCO using RetinaNet.}
    \label{fig:SOTA}
\end{figure*}

\textbf{Uncertainty Re-weighting.} To ensure that the instance uncertainty is consistent with the image uncertainty, we assemble the image classification scores for all classes to a score vector $w_i$ and re-weight the instance uncertainty as
\begin{equation}
    \label{eq_discrepancy_with_mil_weight}
    \tilde{l}_{dis}(x) = \sum_{i}|{w_i}\cdot({\hat y_{i}^{f_1}} - {\hat y_{i}^{f_2}})|,
\end{equation}
where $w_i = \hat{y}_{i}^{cls}$. We then update Eq.~(\ref{eq_max_discrepancy}) to
\begin{equation}
    \label{eq_max_discrepancy_weighted}
     \argmin_{\tilde{\Theta} \setminus \theta_{g}}\tilde{{\cal L}}_{max} = \sum_{x\in {\cal X}_L}\Big(l_{det}(x)+l_{imgcls}(x)\Big) - \sum_{x\in {\cal X}_U} \lambda \cdot \tilde{l}_{dis}(x),
\end{equation}
where $\tilde{\Theta}=\Theta \cup \{\theta_{f_{mil}}\}$.
By optimizing Eq.~(\ref{eq_max_discrepancy_weighted}), the discrepancies of instances with large image classification scores are preferentially estimated, while those with small classification scores are suppressed. 
Similarly, Eq.~(\ref{eq_min_discrepancy}) is updated to
\begin{equation}
    \label{eq_min_discrepancy_weighted}
    \begin{split}
     \argmin_{\theta_g, \theta_{f_{mil}}}\tilde{{\cal L}}_{min} = & {\sum_{x\in {\cal X}_L}\Big(l_{det}(x)+l_{imgcls}(x)\Big)} \\
     & + \sum_{x\in {\cal X}_U} \Big( \lambda \cdot \tilde{l}_{dis}(x) + l_{imgcls}(x) \Big).
     \end{split}
\end{equation}
In Eq.~(\ref{eq_min_discrepancy_weighted}), the image classification loss is applied to the unlabeled set, where the pseudo image labels are estimated using the outputs of the instance classifiers, as
\begin{equation}
    \label{eq_pseudo_image_label}
     y_{c}^{pseudo} = \mathbb{1}\bigg(\max_{i}\Big(\frac{\hat{y}_{i,c}^{f_1} + \hat{y}_{i,c}^{f_2}}{2}\Big), 0.5\bigg),
\end{equation}
where $\mathbb{1}(a, b)$ is a binarization function. When $a > b$, it returns 1; otherwise 0. 
Eq.~(\ref{eq_pseudo_image_label}) is defined based on that instance classifiers can find true instances but are easy to be confused by complex backgrounds. We use the maximum instance score to predict pseudo image labels and leverage MIL to reduce background interference.
%
According to Eqs.~(\ref{eq_mil_score}) and (\ref{eq_mil_loss}), the image classification loss ensures the highlighted instances are representative of the image, \ie, minimizing the image classification loss bridges the gap between the instance uncertainty and image uncertainty. By iteratively optimizing Eqs.~(\ref{eq_max_discrepancy_weighted}) and (\ref{eq_min_discrepancy_weighted}), informative object instances in the same class are statistically highlighted, while the background instances are suppressed. 

\subsection{Informative Image Selection}
\label{sec:topk}
In each learning cycle, after instance uncertainty learning (IUL) and instance uncertainty re-weighting (IUR), we select the most informative images from the unlabeled set by observing the top-$k$ instance uncertainty defined in Eq.~(\ref{eq_max_discrepancy_l_dis}) for each image, where $k$ is a hyper-parameter. This is based on the fact that the noisy instances have been suppressed and the instance uncertainty becomes consistent with the image uncertainty. The selected images are merged into the labeled set for the next learning cycle.

\section{Experiments}
\subsection{Experimental Settings}

\textbf{Datasets}. The $trainval$ sets of PASCAL VOC 2007 and 2012 datasets which contain 5011 and 11540 images are used for training. The VOC 2007 $test$ set is used to evaluate mean average precision (mAP). The MS COCO dataset contains 80 object categories with challenging aspects including dense objects and small objects with occlusion. We use the $train$ set with 117k images for active learning and the $val$ set with 5k images for evaluating AP.


\begin{table*}
\begin{center}
\renewcommand{\arraystretch}{1.3}
\centering
\setlength{\tabcolsep}{1mm}{
\begin{tabular}{|m{12mm}<{\centering}|m{12mm}<{\centering}|
m{12mm}<{\centering}|m{12mm}<{\centering}|m{12mm}<{\centering}|
m{10mm}<{\centering}m{10mm}<{\centering}m{10mm}<{\centering}
m{10mm}<{\centering}m{10mm}<{\centering}m{10mm}<{\centering}
m{10mm}<{\centering}|m{13.5mm}<{\centering}|}
    \hline
	\multicolumn{2}{|c}{Training} & \multicolumn{3}{|c}{Sample Selection} & \multicolumn{8}{|c|}{mAP (\%) on Proportion (\%) of Labeled Images}\\
	\hline
	IUL & IUR & Rand. & Max Unc. & Mean Unc. & 5.0 & 7.5 & 10.0 & 12.5 & 15.0 & 17.5 & 20.0 & 100.0 \\
	\hline\hline
    & & \checkmark & & &            28.31 & 49.42 & 56.03 & 59.81 & 64.02 & 65.95 & 67.09 & \multirow{4}{*}{77.28} \\
	\cline{1-12}
    \checkmark & & \checkmark & & & 30.09 & 49.17 & 55.64 & 60.93 & 64.10 & 65.77 & 67.20 & \\
    \checkmark & & & \checkmark & & 30.09 & 49.79 & 58.94 & 63.11 & 65.61 & 67.84 & 69.01 & \\
    \checkmark & & & & \checkmark & 30.09 & 49.74 & 60.60 & 64.29 & 67.13 & 68.76 & 70.06 & \\
    \cline{1-13}
    & \checkmark & \checkmark & & & 47.18 & 57.12 & 60.68 & 63.72 & 66.10 & 67.59 & 68.48 & \multirow{3}{*}{\bf78.37} \\
    & \checkmark & & \checkmark & & 47.18 & 57.58 & 61.74 & 64.58 & 66.98 & 68.79 & 70.33 & \\
    & \checkmark & & & \checkmark & \bf47.18 & \bf58.03 & \bf63.98 & \bf66.58 & \bf69.57 & \bf70.96 & \bf72.03 & \\
    \hline
\end{tabular}}
\end{center}
\caption{Module ablation on PASCAL VOC. The first line shows the result of the baseline method with random image selection. ``Max Unc.'' and ``Mean Unc.'' respectively denote that the image uncertainty is represented by the maximum and averaged instance uncertainty.}
\label{tab:abl_module}
\end{table*}

\textbf{Active Learning Settings.} We use the RetinaNet~\cite{RetinaNet20} with ResNet-50 and SSD~\cite{SSD16} with VGG-16 as the base detector. 
For RetinaNet, MI-AOD uses 5.0\% of randomly selected images from the training set to initialize the labeled set on PASCAL VOC. In each active learning cycle, it selects 2.5\% images from the rest unlabeled set until the labeled images reach 20.0\% of the training set. 
For the large-scale MS COCO, MI-AOD uses only 2.0\% of randomly selected images from the training set to initialize the labeled set, and then selects 2.0\% images from the rest of the unlabeled set in each cycle until reaching 10.0\% of the training set. 
In each cycle, the model is trained for 26 epochs with the mini-batch size 2 and the learning rate 0.001. After 20 epochs, the learning rate decreases to 0.0001. The momentum and the weight decay are set to 0.9 and 0.0001 respectively.
For SSD, we follow the settings in LL4AL~\cite{LearningLoss19} and CDAL~\cite{CDAL20}, where 1k images in the training set are selected to initialize the labeled set and 1k images are selected in each cycle. The learning rate is 0.001 for the first 240 epochs and reduced to 0.0001 for the last 60 epochs. The mini-batch size is set to 32 which is required by LL4AL.

We compare MI-AOD with random sampling, entropy sampling, Core-set~\cite{CoreSet18}, LL4AL~\cite{LearningLoss19} and CDAL~\cite{CDAL20}. For entropy sampling, we use the averaged instance entropy as the image uncertainty. We repeat all experiments for 5 times and use the mean performance. MI-AOD and other methods share the same random seed and initialization for a fair comparison. $\lambda$ defined in Eqs.~(\ref{eq_max_discrepancy}), (\ref{eq_min_discrepancy}), (\ref{eq_max_discrepancy_weighted}) and (\ref{eq_min_discrepancy_weighted}) is set to 0.5 and $k$ mentioned in Sec.~\ref{sec:topk} is set to 10k.

\subsection{Performance}
\textbf{PASCAL VOC.} In Fig.~\ref{fig:SOTA}, we report the performance of MI-AOD and compare it with state-of-the-art methods on a TITAN V GPU. Using either the RetinaNet~\cite{RetinaNet20} or SSD~\cite{SSD2016} detector, MI-AOD outperforms state-of-the-art methods with large margins. Particularly, it respectively outperforms state-of-the-art methods by 18.08\%, 7.78\%, and 5.19\% when using 5.0\%, 7.5\%, and 10.0\% samples. With 20.0\% samples, MI-AOD achieves 72.27\% detection mAP, which significantly outperforms CDAL by 3.20\%. The improvements validate that MI-AOD can precisely learn instance uncertainty while selecting informative images. When using the SSD detector, MI-AOD outperforms state-of-the-art methods in almost all cycles, demonstrating the general applicability of MI-AOD to object detectors. 

\textbf{MS COCO.} MS COCO is a challenging dataset for more categories, denser objects, and larger scale variation, where MI-AOD also outperforms the compared methods, Fig.~\ref{fig:SOTA}. Particularly, it respectively outperforms Core-set and CDAL by 0.6\%, 0.5\%, and 2.0\%, and 0.6\%, 1.3\%, and 2.6\% when using 2.0\%, 4.0\%, and 10.0\% labeled images.

\begin{table}[!t]
\begin{center}
\renewcommand{\arraystretch}{1.3}
\setlength{\tabcolsep}{1.25mm}{
\begin{tabular}{|m{13mm}<{\centering}|
m{11mm}<{\centering}m{11mm}<{\centering}m{11mm}<{\centering}
m{11mm}<{\centering}m{11mm}<{\centering}|}
\hline
Training & \multicolumn{5}{c|}{mAP (\%) on Proportion (\%) of Labeled Imgs.} \\
\hline
IUL & 2.0 & 4.0 & 6.0 & 8.0 & 10.0 \\
\hline
\hline
& 51.01 & 61.48 & 69.14 & 75.14 & 79.77 \\
\checkmark & \bf58.07 & \bf67.75 & \bf74.91 & \bf78.88 & \bf80.96 \\
\hline
\end{tabular}}
\end{center}
\caption{The effect of IUL for active image classification. Experiments are conducted on CIFAR-10 using the ResNet-18 backbone while the images are randomly selected in all cycles.}
\label{tab:abl_cls}
\end{table}

\begin{table}[!t]
\begin{center}
\renewcommand{\arraystretch}{1.3}
\setlength{\tabcolsep}{0.2mm}{
\begin{tabular}{|m{9mm}<{\centering}|m{5.3mm}<{\centering}|
m{9.1mm}<{\centering}m{9.1mm}<{\centering}m{9.1mm}<{\centering}
m{9.1mm}<{\centering}m{9.1mm}<{\centering}m{9.1mm}<{\centering}
m{9.1mm}<{\centering}|}
\hline
\multirow{2}{*}{$w_i$} & \multirow{2}{*}{Set} & \multicolumn{7}{c|}{mAP (\%) on Proportion (\%) of Labeled Imgs.} \\
\cline{3-9}
& & 5.0 & 7.5 & 10.0 & 12.5 & 15.0 & 17.5 & 20.0 \\
\hline
\hline
1     & $\varnothing$  & 30.09 & 49.17 & 55.64 & 60.93 & 64.10 & 65.77 & 67.20 \\
$\hat{y}_i^{f_1}$ & $\varnothing$ & 31.67 & 50.67 & 55.93 & 60.78 & 64.17 & 66.22 & 67.30 \\
1     & ${\cal X}_L$ & 42.52 & 54.08 & 57.18 & 63.43 & 65.04 & 66.74 & 68.32 \\
$\hat{y}_{i}^{cls}$ & ${\cal X}$   & \bf47.18 & \bf57.12 & \bf60.68 & \bf63.72 & \bf66.10 & \bf67.59 & \bf68.48 \\
\hline
\end{tabular}}
\end{center}
\caption{Ablation study on IUR. ``$w_i$'' is the $w_i$ in Eq.~(\ref{eq_discrepancy_with_mil_weight}). ``Set'' denotes the sample set for IUR. ${\cal X}$ and ${\cal X}_L$ denote the whole sample set and the labeled set, respectively.}
\label{tab:abl_mil}
\end{table}

\subsection{Ablation Study}

\textbf{IUL.} 
As shown in Tab.~\ref{tab:abl_module}, with IUL, the detection performance is improved up to 70.06\% in the last cycle, which outperforms the Random method by 2.97\% (70.06\% $vs.$ 67.09\%). In Tab.~\ref{tab:abl_cls}, IUL also significantly improves the image classification performance with active learning on CIFAR-10. Particularly when using 2.0\% samples, it improves the classification performance by 7.06\% (58.07\% $vs.$ 51.01\%), demonstrating the effectiveness of the discrepancy learning module for instance uncertainty estimation. 

\textbf{IUR.} In Tab.~\ref{tab:abl_module}, IUL achieves comparable performance with the method using the random image selection strategy in the early cycles. This is because there are significant noisy instances that make the instance uncertainty inconsistent with image uncertainty.
After using IUR to re-weight instance uncertainty, the performance at early cycles is improved by 5.04\%$\sim$17.09\% in the first three cycles (row 4 $vs.$ row 1 in Tab.~\ref{tab:abl_mil}). 
In the last cycle, the performance is improved by 1.28\% (68.48\% $vs.$ 67.20\%) in comparison with IUL and 1.39\% in comparison with the Random method (68.48\% $vs.$ 67.09\%). As shown in Tab.~\ref{tab:abl_mil}, image classification score $\hat{y}_{i}^{cls}$ is the best re-weighting metric (row 4 $vs.$ others).
%
Interestingly, when using 100.0\% images for training, the detector with IUR outperforms the detector without IUR by 1.09\% (78.37\% $vs.$ 77.28\%).
These results clearly verify that the IUR module can suppress the interfering instances while highlighting more representative ones, which can indicate informative images for detector training.
%

\begin{table}[!t]
\begin{center}
\renewcommand{\arraystretch}{1.3}
\setlength{\tabcolsep}{0.45mm}{
\begin{tabular}{|m{5.2mm}<{\centering}|m{5.8mm}<{\centering}|
m{9.1mm}<{\centering}m{9.1mm}<{\centering}m{9.1mm}<{\centering}
m{9.1mm}<{\centering}m{9.1mm}<{\centering}m{9.1mm}<{\centering}
m{9.1mm}<{\centering}|}
\hline
\multirow{2}{*}{$\lambda$} & \multirow{2}{*}{$k$} & \multicolumn{7}{c|}{mAP (\%) on Proportion (\%) of Labeled Imgs.} \\
\cline{3-9}
& & 5.0 & 7.5 & 10.0 & 12.5 & 15.0 & 17.5 & 20.0 \\
\hline
\hline
2   & 10k   & 47.18 & 56.94 & 64.44 & 67.70 & 69.58 & 70.67 & 72.12 \\
1   & 10k   & 47.18 & 57.30 & \bf64.93 & 67.40 & 69.63 & 70.53 & 71.62 \\
0.5 & 10k   & 47.18 & \bf58.41 & 64.02 & \bf67.72 & \bf69.79 & \bf71.07 & \bf72.27 \\
0.2 & 10k   & 47.18 & 58.02 & 64.44 & 67.67 & 69.42 & 70.98 & 72.06 \\
\cline{1-9}
0.5 & $N$ & 47.18 & 58.03 & 63.98 & 66.58 & 69.57 & 70.96 & 72.03 \\
0.5 & 10k & 47.18 & 58.41 & \bf64.02 & \bf67.72 & \bf69.79 & \bf71.07 & \bf72.27 \\
0.5 & 100 & 47.18 & \bf58.74 & 63.62 & 67.03 & 68.63 & 70.26 & 71.47 \\
0.5 & 1   & 47.18 & 57.58 & 61.74 & 64.58 & 66.98 & 68.79 & 70.33 \\
\hline
\end{tabular}}
\end{center}
\caption{Performance under different hyper-parameters.}
\label{tab:abl_param}
\end{table}

\begin{table}[!t]
\begin{center}
\renewcommand{\arraystretch}{1.3}
\setlength{\tabcolsep}{0.1mm}{
\begin{tabular}{|m{16mm}<{\centering}|
m{9.3mm}<{\centering}m{9.3mm}<{\centering}m{9.3mm}<{\centering}
m{9.3mm}<{\centering}m{9.3mm}<{\centering}m{9.3mm}<{\centering}
m{9.3mm}<{\centering}|}
\hline
\multirow{2}{*}{Method} & \multicolumn{7}{c|}{Time (h) on Proportion (\%) of Labeled Imgs.} \\
\cline{2-8}
& 5.0 & 7.5 & 10.0 & 12.5 & 15.0 & 17.5 & 20.0 \\
\hline
\hline
Random & 0.77 & 1.12 & 1.45 & 1.78 & 2.12 & 2.45 & 2.78 \\
\cline{1-8}
CDAL~\cite{CDAL20} & 1.18 & 1.50 & 1.87 & 2.19 & 2.68 & \bf2.83 & \bf2.82 \\
MI-AOD & \bf1.03 & \bf1.42 & \bf1.78 & \bf2.18 & \bf2.55 & 2.93 & 3.12 \\
\hline
\end{tabular}}
\end{center}
\caption{{Comparison of time cost on PASCAL VOC.}}
\label{tab:time}
\end{table}

\textbf{Hyper-parameters and Time Cost.} The effects of the regularization factor $\lambda$ defined in Eqs.~(\ref{eq_max_discrepancy}), (\ref{eq_min_discrepancy}), (\ref{eq_max_discrepancy_weighted}) and (\ref{eq_min_discrepancy_weighted}) and the valid instance number $k$ in each image for selection are shown in Tab.~\ref{tab:abl_param}. MI-AOD has the best performance when $\lambda$ is set to 0.5 and $k$ is set to 10k (for $\sim$100k instances/anchors in each image).
Tab.~\ref{tab:time} shows that MI-AOD costs less time at early cycles than CDAL.

\subsection{Model Analysis}
\begin{figure}[t]
    \centering
    \includegraphics[width=1\linewidth]{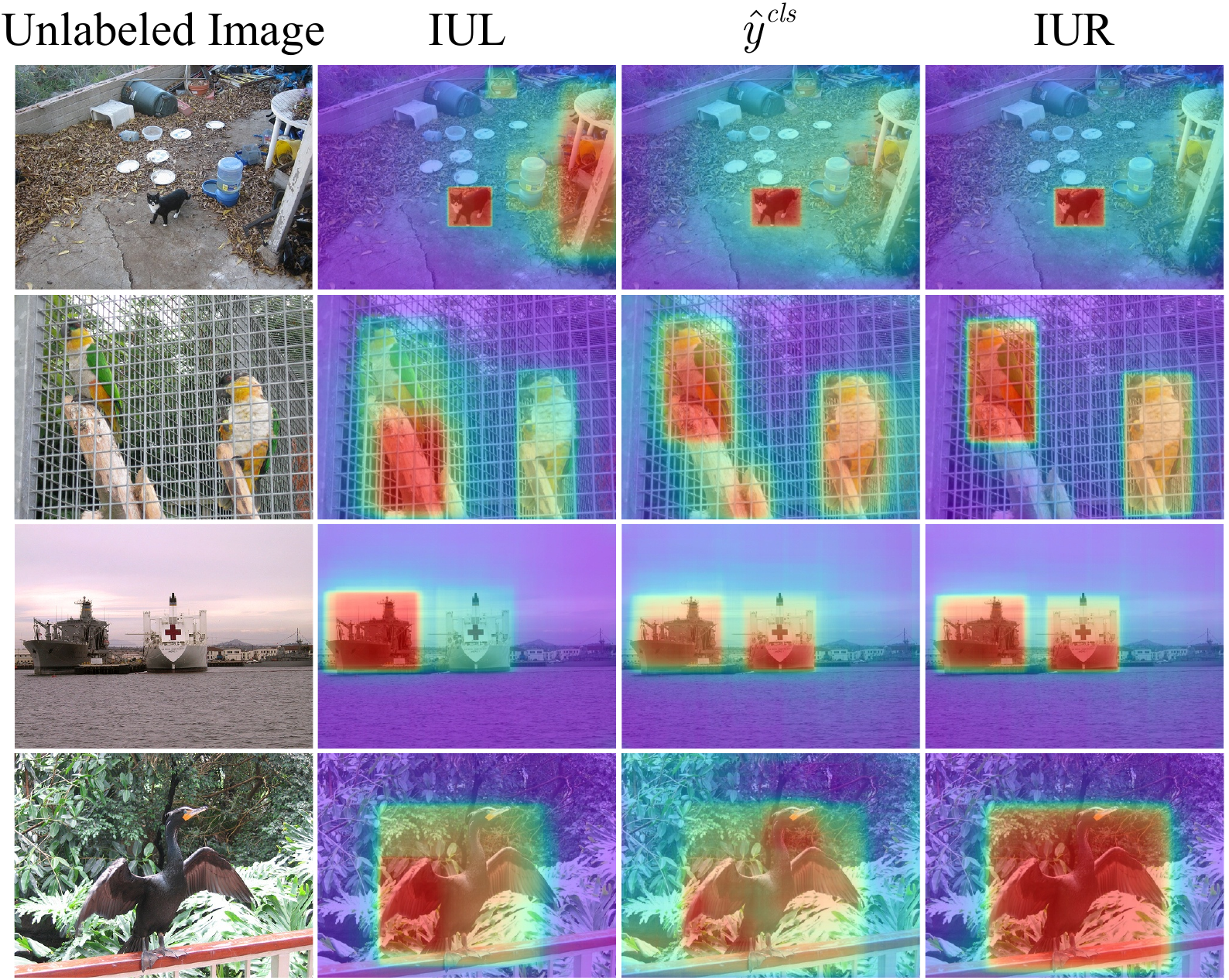}
    \caption{Visualization of learned and re-weighted instance uncertainty and image classification score. (Best viewed in color)
    }
    \label{fig:vis}
\end{figure}

\textbf{Visualization Analysis.} In Fig.~\ref{fig:vis}, we visualize the learned and re-weighted uncertainty and image classification scores of instances. The heatmaps are calculated by summarizing the uncertainty scores of all instances. With only IUL, there exist interference instances from the background (row 1) or around the true positive instance (row 2), and the results tend to miss the true positive instances (row 3) or instance parts (row 4).
MIL can assign high image classification scores to the instances of interesting while suppressing backgrounds. As a result, IUR leverages the image classification scores to re-weight instances towards accurate instance uncertainty prediction.

\begin{figure}[t]
    \centering
    \includegraphics[width=1\linewidth]{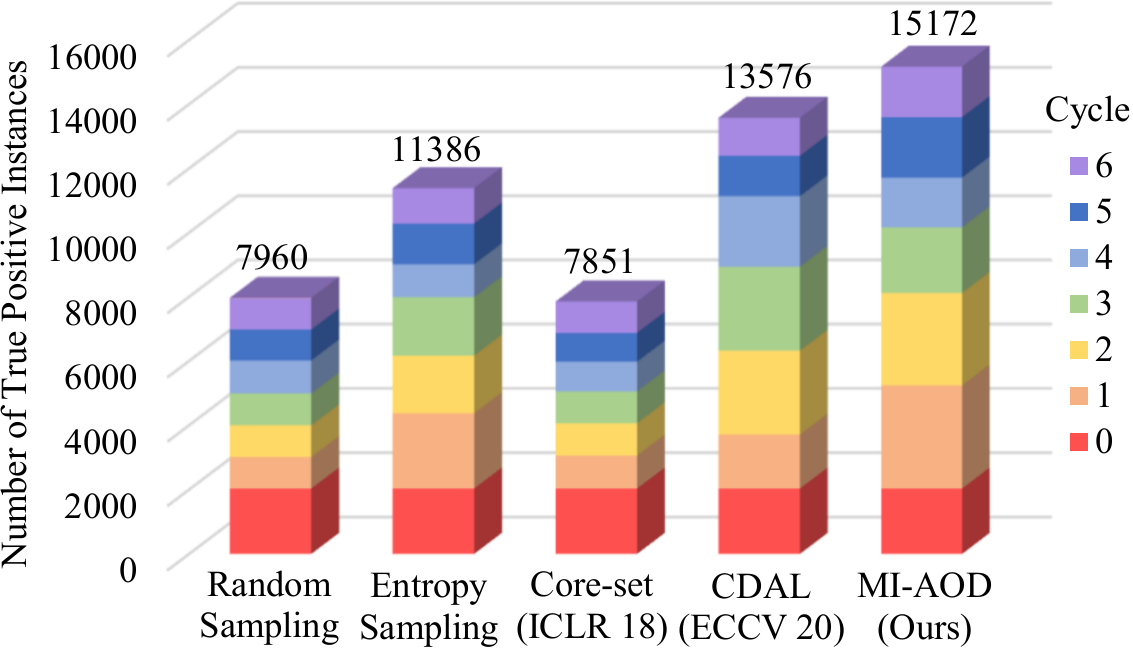}
    \caption{The number of true positive instances selected in each active learning cycle on PASCAL VOC using RetinaNet.}
    \label{fig:stat}
\end{figure}

\textbf{Statistical Analysis.} In Fig.~\ref{fig:stat}, we calculate the number of true positive instances selected in each active learning cycle. It can be seen that MI-AOD significantly hits more true positives in all learning cycles. This shows that the proposed MI-AOD approach can activate true positive objects better while filtering out interfering instances, which facilities selecting informative images for detector training.

\section{Conclusion}
We proposed Multiple Instance Active Object Detection (MI-AOD) to select informative images for detector training by observing instance uncertainty. MI-AOD incorporates a discrepancy learning module, which leverages adversarial instance classifiers to learn the uncertainty of unlabeled instances. MI-AOD treats the unlabeled images as instance bags and estimates the image uncertainty by re-weighting instances in a multiple instance learning (MIL) fashion. Iterative instance uncertainty learning and re-weighting facilitate suppressing noisy instances, towards selecting informative images for detector training. Experiments on large-scale datasets have validated the superiority of MI-AOD, in striking contrast with state-of-the-art methods. 
MI-AOD sets a solid baseline for active object detection.

\textbf{Acknowledgement.} This work was supported by Natural Science Foundation of China (NSFC) under Grant 62006216, 61836012 and 61620106005, the Strategic Priority Research Program of Chinese Academy of Sciences under Grant No. XDA27000000, Post Doctoral Innovative Talent Support Program of China under Grant 119103S304, CAAI-Huawei MindSpore Open Fund and MindSpore deep learning computing framework at \href{https://www.mindspore.cn}{www.mindspore.cn}

{\small
\bibliographystyle{ieee_fullname}
\bibliography{reference}
}

\end{document}